%% file: main_lncs.tex
\documentclass[runningheads]{llncs}

\usepackage[T1]{fontenc}
\usepackage{booktabs}
\usepackage{graphicx}
\usepackage{wrapfig}
\usepackage{multirow}
\usepackage{color}
\usepackage{soul}
\usepackage{amsmath,amssymb}
\usepackage{algorithm}
\usepackage{algorithmic}
\usepackage{adjustbox}
\usepackage{cite}
\usepackage[hidelinks]{hyperref}
\begin{document}

\title{A CPU-Only Framework for Mining Collectively-Behaving Bots in MMORPGs}
\titlerunning{CPU-Only Mining of Collectively-Behaving Bots in MMORPGs}

\author{Hyunsoo Kim \and Jaeman Son \and Sangdeok Seo \and Ilhan Bang \and Man Jae Cho}
\authorrunning{H. Kim et al.}
\institute{NC, Republic of Korea\\
\email{\{aitch25,jaemanson,kuross,ilhanbang,asiasi11\}@ncsoft.com}}

\maketitle

\begin{abstract}
\sloppy
MMORPGs continually face automated programs (bots) that disrupt gameplay balance and in-game economies. An unsupervised Transformer framework can mine collectively behaving bot groups, but its multi-GPU training is costly in live operations. We show that such GPU-trained sequence models are not strictly necessary for this task. Our CPU pipeline pairs zone embeddings learned from movement context, which place teleport-linked zones close together, with a training-free coordinate kernel, and then applies density clustering. It trains 3--5$\times$ faster than the GPU-based Transformer (56--86$\times$ on the same CPU) while matching its mining quality on two commercial MMORPGs, and has been deployed since late 2024. We also share deployment lessons on how to set the clustering threshold and when to retrain.

\end{abstract}

\input{./subtexts/1_Introduction.tex}
\input{./subtexts/2_Background.tex}
\input{./subtexts/3_ProposedApproaches.tex}
\input{./subtexts/4_ExperimentalResult.tex}
\input{./subtexts/5_Conclusion.tex}

\vspace{\baselineskip}

\bibliographystyle{splncs04}
\bibliography{sample-base}

\end{document}

%% file: subtexts/1_Introduction.tex
\section{Introduction}
\label{sec: Introduction}

MMORPGs face persistent threats from automated programs (bots), which exploit game systems to farm resources and conduct real-money trading (RMT), disrupting gameplay balance and in-game economies \cite{huhh2008simple,kwon2016crime,lee2018no}. BotTRep~\cite{kim2024framework} mines collectively behaving bot groups without labels. Moreover, as it relies on a single feature, movement trajectories, and a dedicated visualization, it is clear why a target was flagged, so operators can easily verify it before a sanction and when a false positive is reported. However, its multi-GPU Transformer training is costly: live operations require routine retraining, and GPUs must be shared with other detection models across more games. Other deep learning approaches likewise rely on GPU-intensive infrastructure, and often on large labeled datasets.

We propose \textbf{Lightweight BotTRep} (L-BotTRep), a simpler, interpretable, CPU-only redesign of BotTRep \cite{kim2024framework} that solves the same mining task within these constraints. Its key design separates two views that game worlds entangle: Word2Vec \cite{mikolov2013efficient,mikolov2013distributed} zone embeddings learned from movement context, which place teleport-linked zones close, and a training-free RBF kernel \cite{rahimi2007random,rahimi2008weighted} over raw coordinates. L-BotTRep keeps this property, as movement trajectories remain its only input (Fig.~\ref{fig:fig_heatmap_examples}); it is harder to obtain with methods built on composite features~\cite{chen2009game,lee2018no,lee2016you,pao2010game,pu2022unsupervised,qi2022gnn,su2022trajectory,tao2019mvan,zhao2022t}.

Our contributions are twofold: (i)~to our knowledge, the first CPU-only representation for local-coordinate game worlds that pairs learned zone context with a fixed coordinate kernel, solving the mining task of BotTRep at matching quality while training 3--5$\times$ faster than BotTRep on GPUs (56--86$\times$ on the same CPU); and (ii)~lessons, drawn from extensive experiments, on clustering thresholds, retraining, and model choice across two live game services.

We evaluate the proposed framework on real gameplay data from two commercial MMORPGs, anonymized as Game~A and Game~B. The system has been continuously deployed in the live bot analysis operations of a major game publisher since late 2024.

%% file: subtexts/2_Background.tex
\section{Related Work}
\label{sec:related}

\subsubsection{Game Bot Detection.}
Prior work has explored numerous gameplay signals for detecting abnormal users, including status logs, event records, trading histories, and movement trajectories \cite{chen2009game,lee2018no,pao2010game,qi2022gnn,su2022trajectory,tao2019mvan,zhao2022t}. A notable line of research focuses on trajectory-based detection, where earlier studies \cite{chen2009game,pao2010game} relied on handcrafted movement features (path smoothness, detours, turning patterns), while recent approaches \cite{kim2024framework,su2022trajectory,zhao2022t} learn suspicious trajectory structures with deep models. BotTRep \cite{kim2024framework} introduced an unsupervised trajectory representation model suitable for label-free environments, but its Transformer takes over 14 hours to train on 1.5M samples even with two A40 GPUs (Table~\ref{tab:hardware_exp}). Lockstep detectors in social networks~\cite{beutel2013copycatch,cao2014uncovering} often compare users' timestamped actions pairwise, which scales quadratically: DTW-based clustering would take about 54~h on 32 threads for 50,000 Game~A players, far above our daily inference cost (Section~\ref{sec:evaluation}).

\subsubsection{Trajectory Analysis.}
Trajectory mining has been studied extensively across domains \cite{chang2023contrastive,li2018deep,zhou2025blurred}. However, most techniques assume a GPS-like global coordinate system (GCS) with Euclidean proximity. GPS trajectory encoders differ mainly in how they represent trajectories over standard sequence backbones (e.g., dual-feature cells~\cite{chang2023contrastive}, blurred multi-scale patches~\cite{zhou2025blurred}). In contrast, game worlds by design adopt local coordinate systems (LCS) that reuse coordinate ranges across regions and break continuity via teleports. Real-world trajectory algorithms that assume Euclidean space are therefore unsuitable when applied as is, as they would solve the problem in a coordinate system built on different assumptions. For this reason, we propose a model specialized for the trajectory data of game worlds~\cite{kim2024framework}.

%% file: subtexts/3_ProposedApproaches.tex
\section{Proposed Approach}
\label{sec: Proposed Approach}

\subsubsection{Operational Setting.}

In live operations, bots are sanctioned after detection, long-term monitoring, and verification of collectively controlled behavior for real-money trading (RMT). Monitoring is the bottleneck: groups should be found before RMT occurs, yet normal players may also move together temporarily, so candidates must be observed across many sessions, which is infeasible manually at scale. Because false positives carry operational and legal risks, and operators must show sanctioned players who file complaints why they were flagged, we frame the task as \textit{mining} rather than \textit{detection}: the framework surfaces candidate groups for human review instead of automating sanctions.

\subsection{Data Preparation}
\subsubsection{Zones and Coordinate Systems.}

Each grid cell of the game map defines a zone, and players in the same cell at the same time share a zone. Because Game~B uses a coarse tile-based map (up to about $2048\times2048$) and Game~A a fine-grained continuous one (about $600{,}000\times600{,}000$), we use zone sizes of $256\times256$ for Game~B and $120{,}000\times120{,}000$ for Game~A, roughly the extent of a sub-zone or instance dungeon.

Each log entry is $l_i = (l_i^{(x)}, l_i^{(y)}, l_i^{(z)}, l_i^{(c)}, l_i^{(t)})$, where $(x, y, z)$ are spatial coordinates, $c$ is a continent identifier that distinguishes regions reusing overlapping coordinate ranges (e.g., instance dungeons), and $t$ is a timestamp recorded at one-minute intervals. Each element of the raw sequence $L^{p} = \{l_1, \ldots, l_{|L^{p}|}\}$ of player $p$ is mapped to a zone index $\mathbf{z}(l_i) = (\lfloor l_i^{(x)}/g \rfloor, \lfloor l_i^{(y)}/g \rfloor, \lfloor l_i^{(z)}/g \rfloor, l_i^{(c)})$ with a game-dependent zone size $g$ and a coordinate vector $\mathbf{c}(l_i) = (l_i^{(x)}, l_i^{(y)}, l_i^{(z)}, l_i^{(t)})$, yielding the token sequence $S^{p} = \{s_i\} = \{(\mathbf{z}(l_i), \mathbf{c}(l_i))\}$. Zone and coordinate tokens feed the Word2Vec and RBF kernel modules, respectively.

\subsubsection{Training Pairs.}
$S^{p}$ is divided into non-overlapping segments of 32 tokens; such short segments preserve localized regional context, so that both halves of a segment share similar context. Each segment $d_k = (s_1,\ldots,s_{32})$ is randomly assigned to one of two sampling modes and converted into a positive pair $(\mathcal{A}_k, \mathcal{P}_k)$: the \emph{odd--even split} uses the odd- and even-indexed tokens as anchor and positive, and the \emph{half split} uses the first and last 16 tokens. The other pairs in the same mini-batch serve as negatives.

\subsubsection{Downstream Dataset.}
For clustering, each player's full trajectory $\mathcal{T}^{p} = \{(\mathbf{z}(l_i), \mathbf{c}(l_i))\}_{i=1}^{|L^{p}|}$ is divided into consecutive non-overlapping segments of 16 tokens, matching the input length used in training; the last segment keeps its remaining length. This yields $D^{traj} = \{\mathcal{T}^{p_1}, \ldots, \mathcal{T}^{p_N}\}$.

\subsection{A Lightweight Trajectory Representation Model for MMORPGs}
\label{sec:model}

\subsubsection{Design Motivation.}
We encode zone tokens with Word2Vec~\cite{mikolov2013efficient,mikolov2013distributed} and raw coordinates with an RBF kernel mapping~\cite{rahimi2007random,rahimi2008weighted}. Player trajectories are dominated by localized transitions and repetitive flows between neighboring regions rather than complex long-range dependencies, so skip-gram context learning captures them at a fraction of the cost of self-attention. Because Word2Vec learns zone similarity from co-occurrence in movement sequences rather than from coordinates, zones linked by frequent teleports also become close, and mean-pooled RBF features compare point distributions without assuming path continuity.

\subsubsection{Training Model.}

\begin{figure*}[t]
\centering
\newsavebox{\figA}\sbox{\figA}{\includegraphics[height=100pt]{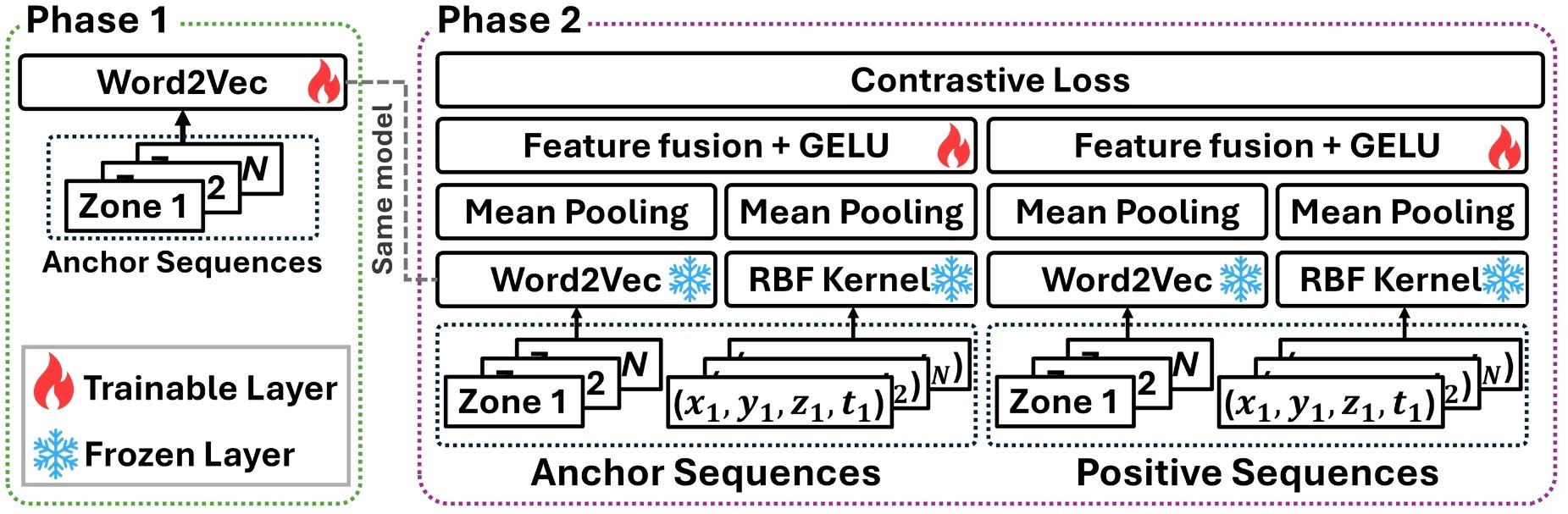}}
\newsavebox{\figB}\sbox{\figB}{\includegraphics[height=100pt]{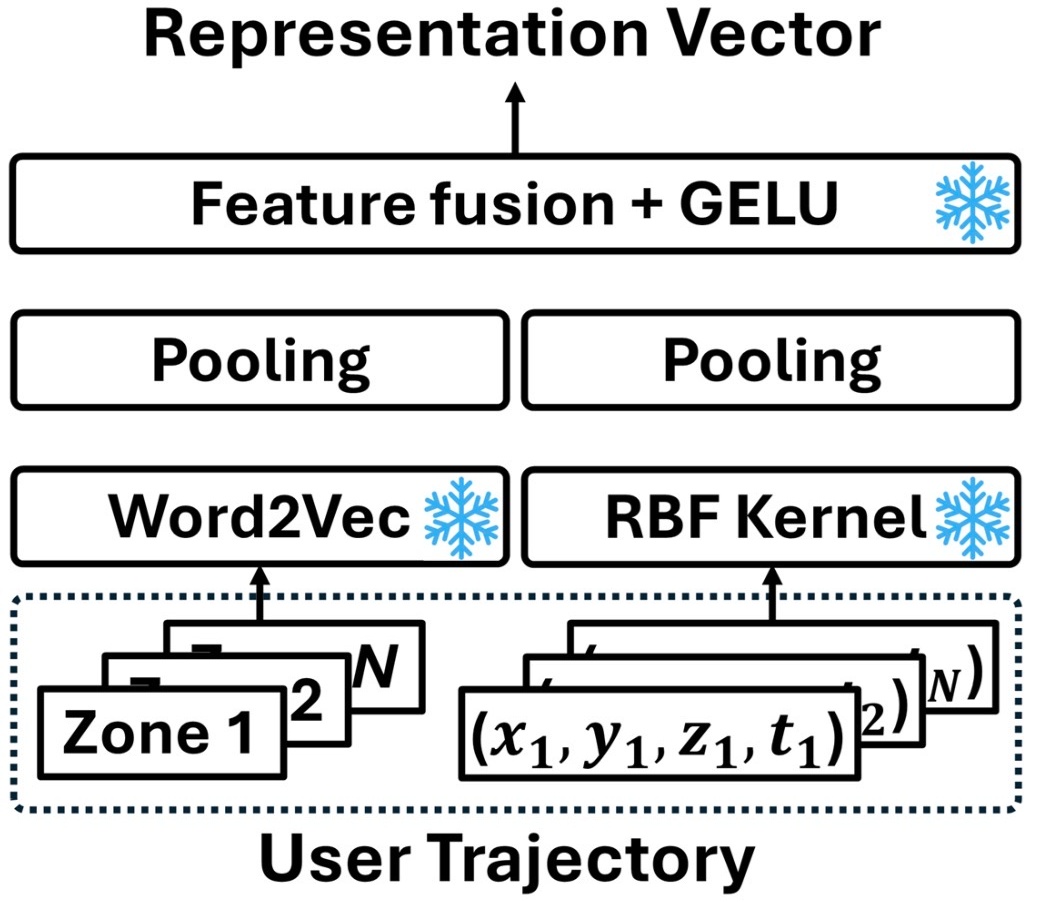}}
\newlength{\figrowh}\setlength{\figrowh}{\dimexpr 100pt*\numexpr\dimexpr 0.98\textwidth\relax\relax/\numexpr\dimexpr\wd\figA+\wd\figB\relax\relax\relax}
\newlength{\figAw}\setlength{\figAw}{\dimexpr\wd\figA*\numexpr\figrowh\relax/\numexpr\dimexpr 100pt\relax\relax\relax}
\newlength{\figBw}\setlength{\figBw}{\dimexpr\wd\figB*\numexpr\figrowh\relax/\numexpr\dimexpr 100pt\relax\relax\relax}
\begin{minipage}[t]{\figAw}
\centering
\includegraphics[height=\figrowh]{./figs/fig_train.jpeg}
\vskip -6pt
\caption{Overview of the proposed framework (training, Phases 1 and 2): Word2Vec and RBF-kernel embeddings are fused through a linear projection layer and optimized with contrastive learning.}
\label{fig: fig_framework}
\end{minipage}\hfill
\begin{minipage}[t]{\figBw}
\centering
\includegraphics[height=\figrowh]{./figs/fig_predict.jpeg}
\vskip -6pt
\caption{\raggedright Inference with the frozen encoder.}
\label{fig:fig_predict}
\end{minipage}
\vskip -2pt
\end{figure*}

As shown in Fig.~\ref{fig: fig_framework}, the proposed model consists of two modules optimized through contrastive learning.

\paragraph{Phase 1.} Word2Vec is pretrained on zone sequences $\mathbf{z}(l_i)$ extracted from real player trajectories to learn contextual relationships among zones.

\paragraph{Phase 2.} A contrastive model is trained on positive pairs $(\mathcal{A}_k, \mathcal{P}_k)$ to pull the two views of a segment together and push apart views of other segments in the mini-batch. It comprises the pretrained Word2Vec and RBF kernel modules, a fusion module (linear projection and GELU), and the NT-Xent loss.

Formally, each zone token $\mathbf{z}(l_i)$ is embedded by the pretrained Word2Vec encoder,
$f_{\mathrm{W2V}}(\mathbf{z}(l_i)) \in \mathbb{R}^D$, which outputs a $D$-dimensional vector for every zone index.
Each coordinate token is represented as $\mathbf{c}(l_i)
=
\left(
l_i^{(x)},
l_i^{(y)},
l_i^{(z)},
l_i^{(t)}
\right)$\footnote{For Game~B, which provides two-dimensional coordinates, $\mathbf{c}(l_i)=(l_i^{(x)}, l_i^{(y)}, l_i^{(t)})$ and $\Omega \in \mathbb{R}^{D\times 3}$.}, and is mapped through random Fourier features (RFF)~\cite{rahimi2007random}, which approximate the RBF kernel
$k(\mathbf{c},\mathbf{c}')=\exp\left(-\gamma\|\mathbf{c}-\mathbf{c}'\|^2\right)$:
\[
\phi_{\mathrm{RBF}}(\mathbf{c}(l_i))
=
\sqrt{2/D}\,
\cos\left(
\Omega\,\mathbf{c}(l_i) + \boldsymbol{\beta}
\right),
\]
where each row of $\Omega \in \mathbb{R}^{D\times 4}$ is drawn from a Gaussian with covariance $2\gamma I$ and
$\boldsymbol{\beta} \in \mathbb{R}^{D}$ is drawn uniformly from $[0, 2\pi]$.
$\Omega$ and $\boldsymbol{\beta}$ are sampled once and kept fixed, so this mapping requires no training,
and $\phi_{\mathrm{RBF}}(\mathbf{c})^{\top}\phi_{\mathrm{RBF}}(\mathbf{c}') \approx k(\mathbf{c},\mathbf{c}')$.
The bandwidth $\gamma$ controls the spatial sensitivity of the resulting $D$-dimensional embedding $\phi_{\mathrm{RBF}}(\mathbf{c}(l_i)) \in \mathbb{R}^D$.

For a view $v$ of up to 16 tokens (the anchor or positive view $\mathcal{A}_k$ or $\mathcal{P}_k$ in training, or a segment of $\mathcal{T}_p$ at inference), the zone and coordinate representations are obtained by mean pooling over its tokens, $\mathbf{e}_z(v)=|v|^{-1}\sum_{i\in v} f_{\mathrm{W2V}}(\mathbf{z}(l_i))$ and $\mathbf{e}_c(v)=|v|^{-1}\sum_{i\in v}\phi_{\mathrm{RBF}}(\mathbf{c}(l_i))$.

To stabilize the representations, Layer Normalization (LN) \cite{ba2016layer} and learnable linear projections are applied to both modality embeddings. The final fused representation is obtained as $\mathbf{h}(v)=\mathrm{GELU}\left(W_1 \mathrm{LN}(\mathbf{e}_z(v))+W_2 \mathrm{LN}(\mathbf{e}_c(v))+\mathbf{b}\right)$,
where $W_1, W_2 \in \mathbb{R}^{D \times D}$ are learnable projection matrices and $\mathbf{b} \in \mathbb{R}^D$ is a learnable bias vector.

Given a mini-batch of $B$ pairs, the $2B$ fused representations are $\ell_2$-normalized, and the model is optimized with the normalized temperature-scaled cross-entropy (NT-Xent) loss~\cite{chen2020simple}, $L=-\frac{1}{2B}\sum_{m=1}^{2B}\log\frac{\exp(\mathrm{sim}(\mathbf{h}_m,\mathbf{h}_{m^+})/\tau)}{\sum_{n\neq m}\exp(\mathrm{sim}(\mathbf{h}_m,\mathbf{h}_n)/\tau)}$, where $m^+$ is the other view of the same segment, $\mathrm{sim}$ is cosine similarity, and $\tau$ is the temperature; the remaining $2B-2$ views act as negatives.

\subsubsection{Feature Extraction Model.}

After training, all model parameters are frozen, and the network is repurposed as a feature extraction model, as illustrated in Fig.~\ref{fig:fig_predict}. Each 16-token segment of a player's daily trajectory is encoded into $\mathbf{h}(v)$, and the $\ell_2$-normalized segment representations are averaged with weights proportional to segment length. The resulting vector is $\ell_2$-normalized again and used for downstream clustering of collectively behaving groups.

\subsection{Mining Collectively-Behaving Groups}
\label{sec:clustering}

After extracting trajectory representations, we identify collectively behaving groups with DBSCAN~\cite{ester1996density}: such groups form dense clusters owing to their highly similar long-term movements, whereas normal players are more diverse and fall into noise. Based on discussions with four game bot detection experts and a legal advisor, we regard long-term coordinated behavior among four or more characters as bot-suspicious, since a single player can hardly control four or more characters simultaneously over extended periods. We therefore set \texttt{min\_samples}=4 and determine $\varepsilon$ adaptively from the distribution of 4-nearest-neighbor distances~\cite{schubert2017dbscan} as $\varepsilon=\mathrm{quantile}_q(4\text{-NN}_{\mathrm{dist}}(D^{traj}))$ with \(q \in [0.05,0.15]\).

%% file: subtexts/4_ExperimentalResult.tex
\section{Experiments and Discussion}
\label{sec: Experiments}

\subsection{Datasets}

Both games use logs from February 8--9, 2024 for training and February 10--11, 2024 for evaluation. For Game~A, training data contain 51,282 players and 1,163,754 trajectory sequences after preprocessing, and evaluation data contain 44,469 players. For Game~B, training data contain 46,366 players and 1,662,110 sequences, and evaluation data contain 45,720 players. All results below come from a rigorous offline evaluation on historical production logs, with the same pipeline and detection logic as in live operation. It covers the full daily player population of every evaluated server, and thus reproduces the important conditions of intended use.

\subsection{Model Setup}
\label{sec:setup}

\subsubsection{L-BotTRep.}
Word2Vec uses 256-dimensional skip-gram embeddings with a context window of 10 and 10 negative samples, trained for 10 epochs. The fusion layer (model and hidden dimensions of 256) is trained for 100 epochs with AdamW (learning rate $10^{-5}$, weight decay $10^{-2}$, batch size 512) and the NT-Xent loss with temperature $\tau=0.1$. We explore $\gamma \in [10^{-7}, 10^{-5}]$ and report the best per game (Table~\ref{tab:detection_performance}; training times use $\gamma = 10^{-5}$). Word2Vec windows of 5 and 15 change TAJ \texttt{pos.} (Section~\ref{sec:evaluation}) by ${<}0.006$; the triplet loss lowers it by 0.02--0.06 (Game~A) and 0.10--0.16 (Game~B) but improves Acc Info by ${\le}0.10$; we keep NT-Xent, as TAJ directly measures co-movement. Removing the RBF branch (Word2Vec only) lowers TAJ \texttt{pos.} at $q=0.05$ from 0.603 to 0.164 (Game~A) and from 0.423 to 0.233 (Game~B) and worsens Acc Info (1.16$\to$6.38 and 1.22$\to$1.78), as distinct groups merge into larger clusters. An RBF-only variant is excluded by design: in LCS game worlds, raw coordinates cannot separate dungeons that share a coordinate range, so it would group players in different dungeons whose paths coincide, a risk that grows as new content reuses coordinate ranges.

\vspace{-8pt}\subsubsection{BotTRep.}
We use our in-house BotTRep variant, which is the one run in operation. It keeps the architecture and hyperparameters of the best-performing configuration of the original study~\cite{kim2024framework}, a Transformer~\cite{vaswani2017attention} with \(d_{\mathrm{model}}=256\), \(d_{\mathrm{hid}}=1024\), \(n_{\mathrm{head}}=16\), and \(n_{\mathrm{layer}}=8\), and is trained for 70 epochs as in that study. It differs in two respects, both inherited by L-BotTRep: the original triplet objective~\cite{schroff2015facenet} is replaced with the NT-Xent loss~\cite{chen2020simple}, and representations are extracted by pooling over 16-token segments, excluding padding tokens. We use a batch size of 512, the largest that fit the low-end GPUs.

\vspace{-8pt}\subsubsection{Sequence Autoencoder Baseline.}

As no unsupervised method other than BotTRep targets this task, we compare against a generic sequence autoencoder (AE): a bidirectional GRU~\cite{cho2014learning} autoencoder (GRU-AE), which outperformed LSTM- and Transformer-based autoencoders in our preliminary experiments. It takes the same fine-grained grid-cell and time tokens as BotTRep, is trained for 70 epochs, and mean-pools its outputs over non-padding tokens.

\vspace{-8pt}\subsubsection{GPS-Trajectory Encoders.}
To test whether GPS-trajectory encoders transfer despite the GCS assumption (Section~\ref{sec:related}), we apply TrajCL~\cite{chang2023contrastive}, a contrastive model over raw coordinates and grid-cell embeddings, and BLUE~\cite{zhou2025blurred}, an autoencoder over progressively blurred multi-scale patches of timestamped points, directly to the in-game $(x,y)$ coordinates of the same training days, split into windows of at most 200 and 256 points, respectively. Both are ported from the authors' official public code, keeping the original architectures, input features, and hyperparameters, including 20 and 30 training epochs; both were already sufficiently trained at these epochs, as training for 70 epochs brought no notable gain. Their grid and blurring cell sizes are set to the TAJ grid, and window embeddings are averaged into one vector per player-day. All models then use the same DBSCAN configuration (Section~\ref{sec:clustering}).

\subsection{Experimental Evaluation}
\label{sec:evaluation}

\begin{table}[t]
\centering
\caption{
Training time comparison across hardware environments and data sizes (number of training samples).
L-, M-, and H- denote low-, medium-, and high-end tiers. Training times of BotTRep and GRU-AE are measured for 70 epochs, and those of TrajCL and BLUE for their original 20 and 30 epochs, all at batch size 512; a dash (--) denotes out of GPU memory; GRU-AE on L and BLUE on L and M could not be trained (out of GPU memory) and are omitted; the best and second-best Full times are in bold and underline. For TrajCL and BLUE, data size counts 200- and 256-point windows, repeated to reach the size. L-BotTRep runs on the L CPU. L: Intel Xeon Silver 4210, 126\,GB RAM, 4$\times$2080Ti (11\,GB each); M: Intel Xeon Silver 4514Y, 2$\times$L4 (24\,GB each); H: AMD EPYC 7302, 4$\times$A40 (48\,GB each; two used).
}
\vskip -10pt
\label{tab:hardware_exp}
\scriptsize
\setlength{\tabcolsep}{1.5pt}
\renewcommand{\arraystretch}{0.9}
\begin{adjustbox}{max width=\columnwidth}
\begin{tabular}{rrrr rrrr rr rrr r}
\toprule
& \multicolumn{3}{c}{\textbf{L-BotTRep}} & \multicolumn{4}{c}{\textbf{BotTRep}} & \multicolumn{2}{c}{\textbf{GRU-AE}} & \multicolumn{3}{c}{\textbf{TrajCL}} & \multicolumn{1}{c}{\textbf{BLUE}} \\
\cmidrule(lr){2-4}\cmidrule(lr){5-8}\cmidrule(lr){9-10}\cmidrule(lr){11-13}\cmidrule(lr){14-14}
\textbf{Size} & W2V & RBF & Full & CPU & L & M & H & M & H & L & M & H & H \\
\midrule
100K & 43s & 1m & \underline{21m} & 19h10m & 2h51m & 1h17m & 1h1m & 1h49m & 1h13m & 33m & 23m & \textbf{20m} & 50m \\
500K & 3m & 6m & \textbf{1h8m} & 3d23h & 15h32m & 6h44m & 5h12m & -- & 9h28m & 2h44m & 1h57m & \underline{1h40m} & 4h11m \\
1M & 6m & 13m & \textbf{2h14m} & 8d0h & 1d8h & 13h55m & 10h59m & -- & 22h17m & 5h25m & 3h54m & \underline{3h27m} & 8h24m \\
1.5M & 8m & 19m & \textbf{3h27m} & 11d7h & 2d2h & 18h19m & 14h27m & -- & 1d9h & 8h9m & 5h50m & \underline{5h10m} & 12h36m \\
\bottomrule
\end{tabular}
\end{adjustbox}
\vskip -5pt
\end{table}

\subsubsection{Resource Efficiency Analysis.}

\vspace{-4pt}\paragraph{Setup.}
To assess deployability on commodity servers, we measured end-to-end training time on subsets of 100K, 500K, 1M, and 1.5M samples drawn from the two-day Game~B dataset on the systems in Table~\ref{tab:hardware_exp}; the low-end system costs approximately USD~10,000.

\vspace{-4pt}\paragraph{Results.}
On the low-end CPU, L-BotTRep trained in 3h 27m for 1.5M samples, whereas BotTRep took over 11 days on the same CPU and 14h 27m on two A40 GPUs. Overall, it was 3--5$\times$ faster than BotTRep on its best GPU setup and 56--86$\times$ faster on the same CPU. Daily inference for about 23,000 players per game takes 6--9~min on an AMD EPYC 7542 CPU, and DBSCAN clustering about 20~s.

\vspace{-6pt}\subsubsection{Mining Performance Evaluation.}

\vspace{-4pt}\paragraph{Time-aware Jaccard similarity.}
TAJ~\cite{kim2024framework} measures temporal trajectory overlap on game-specific grids---\(1500 \times 1500 \times 1500\) for Game~A and \(8 \times 8\) for Game~B, approximately matching the attack range of long-range characters. For each cluster, TAJ is averaged over intra-cluster pairs (\texttt{pos.}) and an equal number of randomly paired trajectories (\texttt{rand.}), and Table~\ref{tab:detection_performance} reports the mean over clusters; coherent clusters show high \texttt{pos.} and low \texttt{rand.} similarity.

As shown in Table~\ref{tab:detection_performance}, the proposed framework matches the mining quality of BotTRep despite its substantially lower computational cost. At $q=0.05$, the best L-BotTRep configuration differs from BotTRep by at most 0.009 in \texttt{pos.} similarity on both games (0.6028 vs.\ 0.6097 on Game~A; 0.4233 vs.\ 0.4319 on Game~B). Both models outperform GRU-AE, clearly so on Game~B, where it (0.3126) trails L-BotTRep by 0.11. TrajCL trails L-BotTRep by 0.046 in \texttt{pos.} on Game~A, slightly below GRU-AE, and by 0.014 on Game~B, where it clearly outperforms GRU-AE. TrajCL benefits from its grid-cell input but still assumes Euclidean proximity (Section~\ref{sec:related}). BLUE reaches only 0.21 and 0.09, below all other models. L-BotTRep attains the best access-information homogeneity on Game~A and is within 0.009 of BotTRep on Game~B (at $\gamma=10^{-6}$), and all models keep \texttt{rand.} similarity close to zero.

\begin{table}[t]
\centering

\caption{
Mining performance of all models at $q=0.05$.
Grp., Samp., TAJ, and Acc Info denote grouped characters (i.e., non-noise samples assigned to clusters), total samples, time-aware Jaccard similarity, and access-info homogeneity, respectively.
\texttt{pos.}: intra-cluster; \texttt{rand.}: random pairs (TAJ) or same-size random groups (Acc Info).
Best and second-best scores are shown in bold and underline, respectively.
}
\vskip -10pt

\footnotesize
\setlength{\tabcolsep}{4pt}
\renewcommand{\arraystretch}{0.9}

\begin{adjustbox}{max width=\columnwidth}
\begin{tabular}{ccccccccc}
\toprule

\multirow{2}{*}{\textbf{Game}}
& \multirow{2}{*}{\textbf{Model}}
& \multicolumn{1}{c}{\textbf{Param}}
& \multirow{2}{*}{\textbf{Grp.}}
& \multirow{2}{*}{\textbf{Samp.}}
& \multicolumn{2}{c}{\textbf{TAJ Sim.}}
& \multicolumn{2}{c}{\textbf{Acc Info}} \\

\cmidrule(lr){3-3}
\cmidrule(lr){6-7}\cmidrule(lr){8-9}

&
&
\boldmath$\gamma$
&
&
& \textbf{\texttt{pos.}}
& \textbf{\texttt{rand.}}
& \textbf{\texttt{pos.}}
& \textbf{\texttt{rand.}} \\

\midrule

\multirow{7}{*}{Game~A}
& \multirow{3}{*}{L-BotTRep}
& $10^{-5}$
& 1,819
& \multirow{4}{*}{44,469}
& 0.5998
& 0.0055
& 1.1834 & 4.93 \\

&
& $10^{-6}$
& 1,839
&
& 0.6013
& 0.0050
& \underline{1.1694} & 4.85 \\

&
& $10^{-7}$
& 1,829
&
& \underline{0.6028}
& 0.0049
& \textbf{1.1620} & 4.85 \\

& BotTRep
& --
& 1,807
&
& \textbf{0.6097}
& 0.0078
& 1.1833 & 4.79 \\
\cmidrule(lr){2-9}
& TrajCL & -- & 1,812 & \multirow{3}{*}{44,469} & 0.5571 & 0.0057 & 1.2299 & 4.95 \\
& BLUE & -- & 1,583 & & 0.2147 & 0.0053 & 2.7684 & 5.18 \\
& GRU-AE & -- & 1,873 & & 0.5850 & 0.0063 & 1.2198 & 4.88 \\

\midrule

\multirow{7}{*}{Game~B}
& \multirow{3}{*}{L-BotTRep}
& $10^{-5}$
& 1,966
& \multirow{4}{*}{45,720}
& \underline{0.4233}
& 0.0113
& 1.2196 & 5.04 \\

&
& $10^{-6}$
& 1,884
&
& 0.4134
& 0.0116
& \underline{1.1585} & 5.11 \\

&
& $10^{-7}$
& 1,878
&
& 0.4161
& 0.0124
& 1.1955 & 5.31 \\

& BotTRep
& --
& 1,827
&
& \textbf{0.4319}
& 0.0155
& \textbf{1.1496} & 5.10 \\
\cmidrule(lr){2-9}
& TrajCL & -- & 1,950 & \multirow{3}{*}{45,720} & 0.4096 & 0.0139 & 1.2474 & 5.19 \\
& BLUE & -- & 1,820 & & 0.0927 & 0.0078 & 1.4864 & 4.41 \\
& GRU-AE & -- & 2,397 & & 0.3126 & 0.0077 & 1.1689 & 5.83 \\

\bottomrule
\end{tabular}
\end{adjustbox}
\label{tab:detection_performance}
\vskip -2pt
\end{table}

\vspace{-4pt}\paragraph{Access information homogeneity.}
For each cluster, we build a graph whose nodes are players, connected if they share at least one access attribute, and report the number of connected components. A score of 1 indicates a cluster plausibly operated by a single entity; lower is better.
Empirically, clusters of four or more members scoring about 1.2 are typically bot groups; as collective bots may also use different access attributes, we use the score as supporting evidence alongside TAJ.

\vspace{-6pt}\subsubsection{Operational Validation and Heatmap-based Qualitative Analysis.}

\begin{table}[t]
\centering
\caption{
Operational validation precision (confirmed/mined groups) across \(q\).
\mbox{L-BotTRep} uses $\gamma=10^{-7}$ for Game~A and $10^{-5}$ for Game~B.
Best and second-best per game and \(q\) are in bold and underline.
}
\vskip -10pt
\label{tab:human_eval}

\footnotesize
\setlength{\tabcolsep}{4pt}
\renewcommand{\arraystretch}{0.9}

\setlength{\tabcolsep}{2.6pt}
\begin{adjustbox}{max width=\textwidth}
\begin{tabular}{ccccccccccc}
\toprule
& \multicolumn{5}{c}{\textbf{Game~A}} & \multicolumn{5}{c}{\textbf{Game~B}} \\
\cmidrule(lr){2-6}\cmidrule(lr){7-11}
\textbf{$q$} & L-BotTRep & BotTRep & TrajCL & BLUE & GRU-AE & L-BotTRep & BotTRep & TrajCL & BLUE & GRU-AE \\
\midrule
0.05 & \underline{.9106} & \textbf{.9222} & .8966 & .3263 & .8929 & .8514 & \textbf{.8796} & \underline{.8591} & .1390 & .7905 \\
0.10 & .8780 & \textbf{.8824} & .8650 & .2345 & \underline{.8784} & \underline{.8404} & \textbf{.8613} & .8310 & .1343 & .7116 \\
0.15 & \textbf{.8431} & .8208 & \underline{.8241} & .2228 & .8129 & \underline{.8031} & \textbf{.8273} & .7816 & .1356 & .6766 \\
\bottomrule
\end{tabular}
\end{adjustbox}

\vskip -4pt
\end{table}

To measure precision at scale, we validate every mined group against the in-house monitoring criteria that the data analytics team uses to confirm collective bots, rather than manually reviewing sampled groups; the criteria are not disclosed to prevent evasion. As false positives are costly (Section~\ref{sec: Proposed Approach}), we prioritize precision over recall. The criteria combine enforcement history, account-level linkage, and behavioral consistency within the group, and are used neither in training nor in model or hyperparameter selection. In Table~\ref{tab:human_eval}, on Game~A, BotTRep and L-BotTRep attain 0.82--0.92 and differ by at most 0.023; GRU-AE is on par up to \(q=0.10\) but trails L-BotTRep by 0.03 at \(q=0.15\). On Game~B, BotTRep and L-BotTRep attain 0.80--0.88 and stay within 0.028 of each other, whereas GRU-AE reaches 0.68--0.79. TrajCL attains 0.82--0.90 and 0.78--0.86 on Games~A and~B, within 0.022 of L-BotTRep, while BLUE stays at 0.13--0.33.

\begin{figure}[t]
\centering

\makebox[0pt][r]{\scriptsize (a)\hspace{2pt}}\adjustbox{valign=c}{\includegraphics[width=0.97\linewidth]{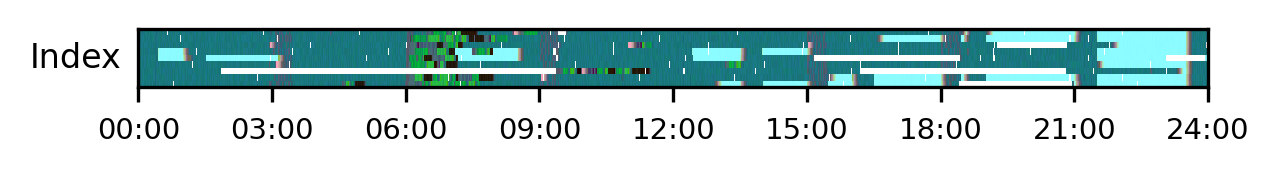}}\par
\vspace{-8pt}
\makebox[0pt][r]{\scriptsize (b)\hspace{2pt}}\adjustbox{valign=c}{\includegraphics[width=0.97\linewidth]{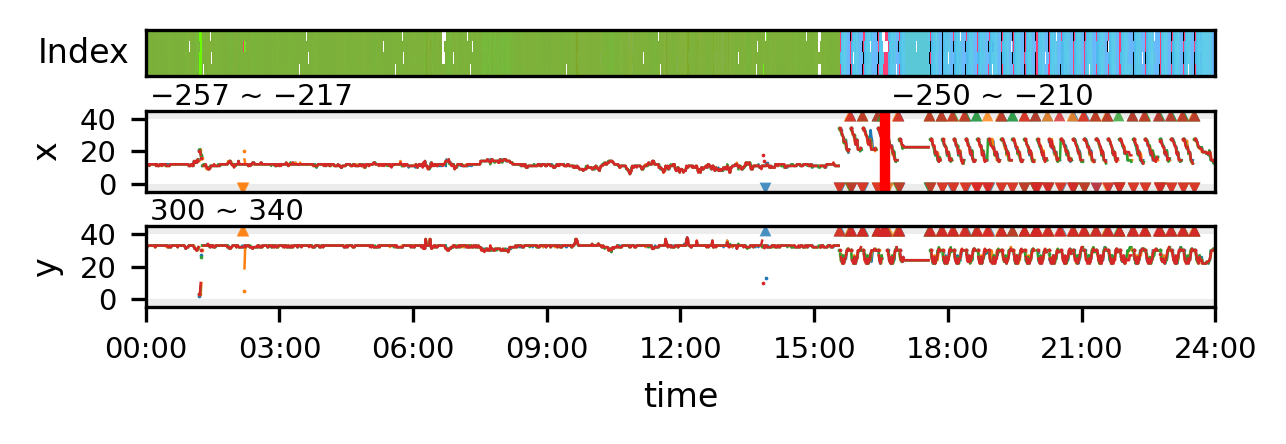}}\par

\vskip -16pt
\caption{
Two Game~A groups (heatmap rows: members; color: location; white: offline). (a) A non-collective group of nine mined at $q=0.15$; (b) a bot group confirmed by per-member \(x\), \(y\) coordinates (one line color per member, so the four lines coincide; 40-cell windows, shifted at red lines; $\blacktriangle\blacktriangledown$: values outside the window).}
\label{fig:fig_heatmap_examples}
\vskip -4pt
\end{figure}

For qualitative analysis, we also inspect trajectory heatmaps of the mined groups. Heatmaps (Fig.~\ref{fig:fig_heatmap_examples}) plot players against the time of day, with locations mapped to RGB colors~\cite{kim2024framework} and offline periods in white. Groups that do not move collectively are often visible at a glance: in the group of Fig.~\ref{fig:fig_heatmap_examples}(a), mined at \(q=0.15\), members play nearly around the clock, unlike typical players, yet differ in color and offline periods. Reviewers can thus screen out such a group in seconds: it shows no collective movement, so it is not treated as a collective bot group without other evidence. Fig.~\ref{fig:fig_heatmap_examples}(b) shows a group that can be regarded as bots; in its first half, however, the members repeat the same color, so it must be checked whether their trajectories actually overlap or whether they are ordinary players hunting repeatedly in one area. This ambiguity is greater in Game~B, whose built-in auto-hunt mode keeps a character roaming near one spot for hours without changing maps (players in a timed dungeon return only when its time ends); there, long co-location alone is not evidence, and reviewers confirm bots only after jointly examining login periods, movement patterns, and other evidence. In such cases, we additionally plot each character's \(x\) and \(y\) coordinates over time, a verification view introduced in this work, to confirm that the movement is truly collective: here the four members' traces overlap throughout the day, including a repeated route after 15:30.

\section{Lessons from Deployment}
\label{sec:lessons}

\subsubsection{$q$ is an operational control parameter.}

Raising \(q\) from 0.05 to 0.15 trades purity for coverage: L-BotTRep groups about three times as many characters (1,829 to 5,455 on Game~A; 1,966 to 5,957 on Game~B), while TAJ \texttt{pos.} drops from 0.60 to 0.51 and from 0.42 to 0.32, Acc Info stays within 1.16--1.22 (random groups: 4.85--5.68), and precision remains above 0.80 (Table~\ref{tab:human_eval}). Although a larger \(q\) widens the candidate pool, in operation we keep \(q=0.05\) in almost all cases, as handling false positives and wrongful sanctions costs more review effort than the extra coverage returns.

\vspace{-8pt}\subsubsection{Weekly retraining scales across games.}
New content ships on the weekly maintenance day, in weeks that cannot be foreseen, and the maintenance days of different games often coincide. We therefore retrain L-BotTRep on every maintenance day without a prior performance check, and more often right after a content update to keep the mined groups stable. Even when several games coincide, this causes no resource contention, as L-BotTRep retrains on a CPU in hours.

\vspace{-8pt}\subsubsection{Generic trajectory encoders do not transfer.}
Established trajectory encoders, which assume Euclidean continuity that coordinate reuse and teleports break, did not replace game-specific modeling: BLUE fails to form coherent groups (TAJ \texttt{pos.} 0.21/0.09), TrajCL comes close but needs GPUs and trails on Game~A, and the autoencoder exceeds low- and mid-tier GPU memory at scale. What made L-BotTRep work was pairing zone context with a coordinate kernel; zone context alone merges distinct groups (Section~\ref{sec:setup}).
\vspace{-8pt}\subsubsection{Co-movement is a lead, and trades expose collectors.}
Bot farms hunt together for efficiency and sell the loot to fund RMT. In operation, each mined group starts a review: reviewers select targets with the heatmap and coordinate views (Fig.~\ref{fig:fig_heatmap_examples}) and weigh illegal-program detections, trade patterns, payments, and content consumption, among which cross-checking abnormal trades has been particularly effective. In our offline data, member accounts often bought items within 30 days at prices flagged as abnormally high by in-house trade monitoring, a common way to hand over in-game currency; among L-BotTRep groups with such purchases, the account receiving the most was paid by two or more member accounts in 79--81\% of groups in Game~A and 70--75\% in Game~B, versus 1--2\% and 17--18\% for random groups of the same size and server ($q=0.05$--$0.15$).

\vspace{-8pt}\subsubsection{Two coordinate archetypes, model choice by stakes.}
The framework now runs on seven games through nine pipelines, as regional services with different content are modeled separately. Their worlds all have coordinate systems similar to that of Game~A or Game~B, so the pipeline transferred almost unchanged. L-BotTRep runs five of the nine; our in-house BotTRep variant runs the other four, newly launched games and games with heavy enforcement, where a marginal precision gain (Table~\ref{tab:human_eval}) is worth its GPU cost.

%% file: subtexts/5_Conclusion.tex
\vspace{-8pt}\section{Conclusion}
\label{sec:Conclusion}

We showed that collective bot mining is possible without a GPU-trained Transformer: interpretable Word2Vec and RBF trajectory features with clustering match BotTRep on a CPU at much lower training cost. For practitioners, zone context with a fixed coordinate kernel sufficed without GPUs, whereas generic trajectory encoders needed GPUs or lost quality.